\documentclass[10pt,journal,compsoc]{IEEEtran}

\long\def\quot#1{{``#1''}}

\usepackage{amsmath,amstext}
\usepackage{algorithm,algpseudocode}
\usepackage{amsfonts}
\usepackage{amsthm}

\usepackage{color}
\definecolor{Orange}{rgb}{1,0.5,0}
\definecolor{Red}{rgb}{1,0,0}

\newcommand{\AF}[1]{\textsf{\textbf{\textcolor{Orange}{[AF: #1]}}}}
\newcommand{\MET}[1]{\textsf{\textbf{\textcolor{Red}{[MET: #1]}}}}

\newcommand{\AFold}[1]{\ignorespaces}
\newcommand{\METold}[1]{\ignorespaces}

\newcommand{\extVersion}[1]{\ignorespaces}

\newcommand{\rev}[1]{{\normalfont\selectfont\color{blue}#1}}

\newcommand{\remove}[1]{{\normalfont\selectfont\color{red}Possibly Remove? #1}}

\renewcommand{\remove}[1]{\ignorespaces} 
\renewcommand{\AF}[1]{\ignorespaces}
\renewcommand{\MET}[1]{\ignorespaces}
\renewcommand{\rev}[1]{#1} 

\usepackage{pbox}

\theoremstyle{definition}
\newtheorem{definition}{Definition}[section]

\usepackage{threeparttable}
\usepackage{standalone}
\usepackage{booktabs, dcolumn}
\usepackage{amssymb}
\usepackage{subcaption}

\usepackage{slashbox}

\theoremstyle{plain}

\theoremstyle{definition}

\ifCLASSOPTIONcompsoc
  \usepackage[nocompress]{cite}
\else
  \usepackage{cite}
\fi
\ifCLASSINFOpdf
\else
\fi
\usepackage{url}

\begin{document}
%
\title{Learning to Teach\\Reinforcement Learning Agents}

%
%
%

%
%
%
%

\author{Anestis~Fachantidis,~\IEEEmembership{}
        Matthew E. Taylor,~\IEEEmembership{}
        and~Ioannis Vlahavas~\IEEEmembership{}
\IEEEcompsocitemizethanks{\IEEEcompsocthanksitem Anestis Fachantidis and Ioannis Vlahavas are with the Department of Informatics, Aristotle University of Thessaloniki, Greece, 5412.\protect\\
E-mail: afa@csd.auth.gr and vlahavas@csd.auth.gr
\IEEEcompsocthanksitem Matthew E. Taylor is with the School of Electrical Engineering and Computer Science, Washington State University, Pullman, WA 99164\protect\\ 
E-mail: taylorm@eecs.wsu.edu}
}

\markboth{}
{Shell \MakeLowercase{\textit{et al.}}: Bare Demo of IEEEtran.cls for IEEE Journals}
%



\maketitle

\begin{abstract}
In this article we study the transfer learning model of action advice under a budget. We focus on reinforcement learning teachers providing action advice to heterogeneous students playing the game of Pac-Man under a limited advice budget. First, we examine several critical factors affecting advice quality in this setting, such as the average performance of the teacher, its variance and the importance of reward discounting in advising. The experiments show the non-trivial importance of the coefficient of variation (CV) as a statistic for choosing policies that generate advice. The CV statistic relates variance to the corresponding mean. Second, the article studies policy learning for distributing advice under a budget. Whereas most methods in the relevant literature rely on heuristics for advice distribution we formulate the problem as a learning one and propose a novel RL algorithm capable of learning when to advise, adapting to the student and the task at hand. Furthermore, we argue that learning to advise under a budget is an instance of a more generic learning problem: Constrained Exploitation Reinforcement Learning.
\end{abstract}


%
\IEEEpeerreviewmaketitle

\section{Introduction}

In the reinforcement learning framework~\cite{sutton98}, data efficient approaches are especially important for real \AF{\#25} world and commercial applications, such as robotics\AF{\#21 removed video games..}. In such domains extensive interaction with the environment needs time and can be costly.

One data efficient approach for RL is \emph{transfer learning} (TL)~\cite{JMLR09-taylor}. Typically, when an RL agent leverages TL, it uses knowledge acquired in one or more (\emph{source}) tasks to speed up its learning in a more complex (\emph{target}) task. Most realistic TL settings require transfer of knowledge between different tasks or heterogeneous agents that can be vastly different from each other (e.g., humans and software agents).

Transferring between heterogeneous agents is often challenging since most methodologies involve exploiting the agents' structural similarity to transfer knowledge between tasks. As an example, TL can be applied between two similar RL agents, which both use the same function approximation method, by transferring their learned parameters. In such a case, a Q-Value transfer solution could be used, combined with an algorithm constructing mappings between the state variables of the two tasks. 

Whereas solutions for extracting similarity between tasks have been extensively studied in the \rev{past \cite{JMLR09-taylor,lazaricsurvey}\AF{\#24}, the} main problem of transferring between very dissimilar agents (e.g., humans and software agents) remains.

\rev{Consider for example a game hint system for human players. The game hint system can not directly transfer its internal knowledge to the human player. Moreover, it should transfer knowledge in a limited and prioritized way since the attention span of humans is limited.}\AF{\#21}

The only prominent knowledge transfer unit between all agents (software, physical or biological) is action. Action suggestion (advice) can be understood by very different agents. However, even when transferring using advice, four problems arise:
\begin{enumerate}
\item Decide what to advise (production of advice)
\item Decide when to advise (distribution of advice), especially when using a limited advice budget 
\item Determine a common action language in order to appropriately express the advice between heterogeneous agents
\item Communicate the advice effectively, ensuring its timely and noiseless reception 
\end{enumerate}


This article focuses on the first two problems---those of deciding when and what to advise under a budget. Moreover, we use the game of Pac-Man to test our methods' effectiveness in a complex domain. 

Whereas works such as~\cite{zhan_online_2015} provide a formal understanding of RL students receiving advice and the implications on the \emph{student's} learning process (e.g. convergence properties) and papers like~\cite{zimmer} and~\cite{torrey13} provide practical methods for a teacher to advise agents, \rev{this work attempts a new learning formulation of the problem and proposes a novel learning algorithm based on it. }\AF{\#1,\#15  BEFORE IT WAS: this work is the first attempt for a formal understanding of the \emph{teacher} agent, which also results in a novel learning model for advising.} We identify and exploit the similarities of the advising under a budget (AuB) problem to the classic exploration-exploitation problem in RL \rev{and identify a} \AF{\#1,\#15  BEFORE IT WAS:by proposing a new} sub-class of reinforcement learning problems: Constrained Exploitation Reinforcement Learning.

Most successful methodologies for AuB require students to inform their teacher for their intended action. This is not a realistic requirement in many real-world TL problems, since it assumes one more communication channel between the student and the \rev{teacher}\AF{\#26 just this?}, thus, it requires some form of structural compliance from the student. \rev{An example of how restrictive is this requirement for real-world applications comes from the game hint example system. The system advises the human player for his next action  in real-time but the human player could never be expected to announce its intended action beforehand. Part of this work's goal is also to alleviate such a prerequisite and propose methods that can also work without such knowledge.}\AF{\#5}  

Specifically, the contributions of this article are:
\begin{itemize}
\item An empirical study on determining an appropriate advising policy in the game of Pac-Man
\item A novel application of average reward reinforcement learning to produce advice
\item \rev{A novel formulation of the learning to advise under budget (AuB) problem as a problem of constrained exploitation RL}\AF{\#1,\#15}  
\item A novel RL algorithm for learning a teaching policy to distribute advice, \rev{able to train faster (lower data complexity) than previous learning approaches and advise even when not having knowledge of the student's intended action}\AF{\#5} 
\end{itemize} 


\section{Background}
\label{sec:background}
This section presents the necessary background to understand the methods proposed in this article. Brief introductions are provided to reinforcement learning and transfer learning, which are then followed by a more detailed discussion of the current advising methodologies. 
\subsection{Reinforcement Learning}
\label{sec:rl}
Reinforcement Learning addresses the problem of how an agent
can learn a behaviour through trial-and-error interactions with a
dynamic environment~\cite{sutton98}. In an RL task the agent, at each
time step, senses the environment's state, $s\in S$, where $S$ is
the finite set of possible states, and selects an action $a\in
A(s)$ to execute, where $A(s)$ is the finite set of possible
actions in state $s$. \rev{The agent receives a reward, $r\in
\mathbb{R}$, and moves to a new state $s'\in S$ according to a transition function, $T$, of the task with $T(s,a,s') = P(s'|s,a)$. }\AF{\#22} The general goal of the agent is to maximize the expected return, where the return, $G$, is defined as some specific function of the reward sequence given also a discounting parameter, $\gamma$. The $\gamma$ parameter, where $0\leq\gamma<1$, controls the importance of short-term rewards over the most long-term ones, discounting the later by powers of factor of $\gamma$. 

The outcome will be an action-value function $Q^{\pi}(s,a)$ which expresses the expected return starting from $s$, taking action $a$, and following after that policy  $\pi: S \rightarrow A$, which dictates
how the agent acts in a certain situation in order to maximize the reward received over time.


\subsection{Transfer Learning and Advising under a Budget}
\label{background2}

Transfer Learning~\cite{JMLR09-taylor} refers to the process of using knowledge that has been acquired in a previously learned task, the \emph{source task}, in order to enhance the learning procedure in a new and more complex task, the \emph{target task}. The more similar these two tasks are, the easier it is to transfer knowledge between them. By similarity, we mean the similarity of their underlying Markov Decision Processes (MDP) that is, the transition and reward functions of the two tasks and also their state and action spaces. 

The type of knowledge that can be transferred between tasks varies among different TL methods, including value functions, entire policies, actions (policy advice) or a set of samples from a source task which can be used by a model-based RL algorithm in a target task.

\remove{Whereas the typical TL problems can be handled by TL algorithms such as TIMBREL~\cite{taylor08b}, the problem of advising agents has significant differences and poses new challenges. In typical TL, agents perform transfer knowledge from one task to another and not between agents. Based on a recent characterization~\cite{zhan_online_2015} typical TL can be considered as off-line TL and advising agents as on-line TL, a distinct category describing problems where a teacher agent represents the source task knowledge and actively assists another agent, the student.} 

Focusing specifically on policy advice under an advice budget constraint, we identify two aspects of the problem, a) learning a policy to produce advice and b) distributing the advice in the most appropriate way, while respecting the advice budget constraint. Most methods in the literature produce advice by greedily using a learned policy for the task in hand~\cite{torrey13,zimmer,zhan_online_2015}. For advice distribution, most methods rely on some form of heuristic function (and not learning) based on which the teacher decides when to advice. Examples of such methods are Importance Advice and Mistake Correcting~\cite{torrey13}. 

The Importance Advice method produces advice by repeatedly querying a learned policy's value function, on each state the student faces, to obtain the best action for that state. Distribution of advice, that is deciding when to advise or not, is determined by a heuristic logical expression of the form $Q_{max}(s,a) - Q_{min}(s,a)>t$ where $t$ is a threshold parameter determining the state-action value gap between the best and the worst action for that state. If this value gap exceeds the threshold value, $t$, the state is considered critical and advice is given. The algorithm continues until the advice budget finishes.

Mistake correcting (MC)~\cite{torrey13} differs from Importance Advising only in presuming knowledge of the student's intended action. Consequently, it validates the Importance Advising criterion only if the student action is wrong, not wasting advice when the student does not need it.

\extVersion{. If this is the case, we can use MC, which gives advice only when the value gap exceeds $t$ \emph{and} the student's intended action is different from the teacher's best action. This way, MC does not waste advice in states where the student would do the correct action anyway, on its own.

Besides the above heuristic-based methods and to the best of the authors' knowledge, there is just one method applying learning for the distribution of advice. 
}

The method presented in~\cite{zimmer} \rev{(Zimmer's method )}\AF{\#26} formulates the teaching problem as an RL one in order to learn an advice distribution policy. The teacher agent has an action set with two actions, $A=\{$advice, no advice$\}$. The teacher's state space is an augmented version of the student's one and is of the form: $s_{teacher} = (s_{student} , a_{student} , b, n_e )$ where $s_{student}$ is the current state vector of the student, $a_{student}$ is the intended action of the student (it is assumed that the student announces the intended action on every step), $b$ the remaining advice budget and $n_e$ the student's training episode number. Moreover, the reward signal is a transformed version of the student's reward with an extra positive reward for the teacher when the student reaches its goal in a small number of steps. We note that this method is tested only on the Mountain Car domain and the reward signal proposed for the teacher is domain-dependent.
 
The policy advice methods~\cite{torrey13},\cite{zimmer},\cite{zhan_online_2015}, presented in this section  will also be used for comparisons in the experiments presented later in this article.\AF{\#33} \extVersion{ and thus appropriate only for domains with similar reward structure to that of the Mountain Car domain.}

\extVersion{Additionally to the advice distribution methods, the impact of a teacher's advice on RL students using a finite budget of advice has been recently studied in~\cite{zhan_online_2015}.}

\begin{figure}[!t]
\centering
\includegraphics[width=0.21\textwidth]{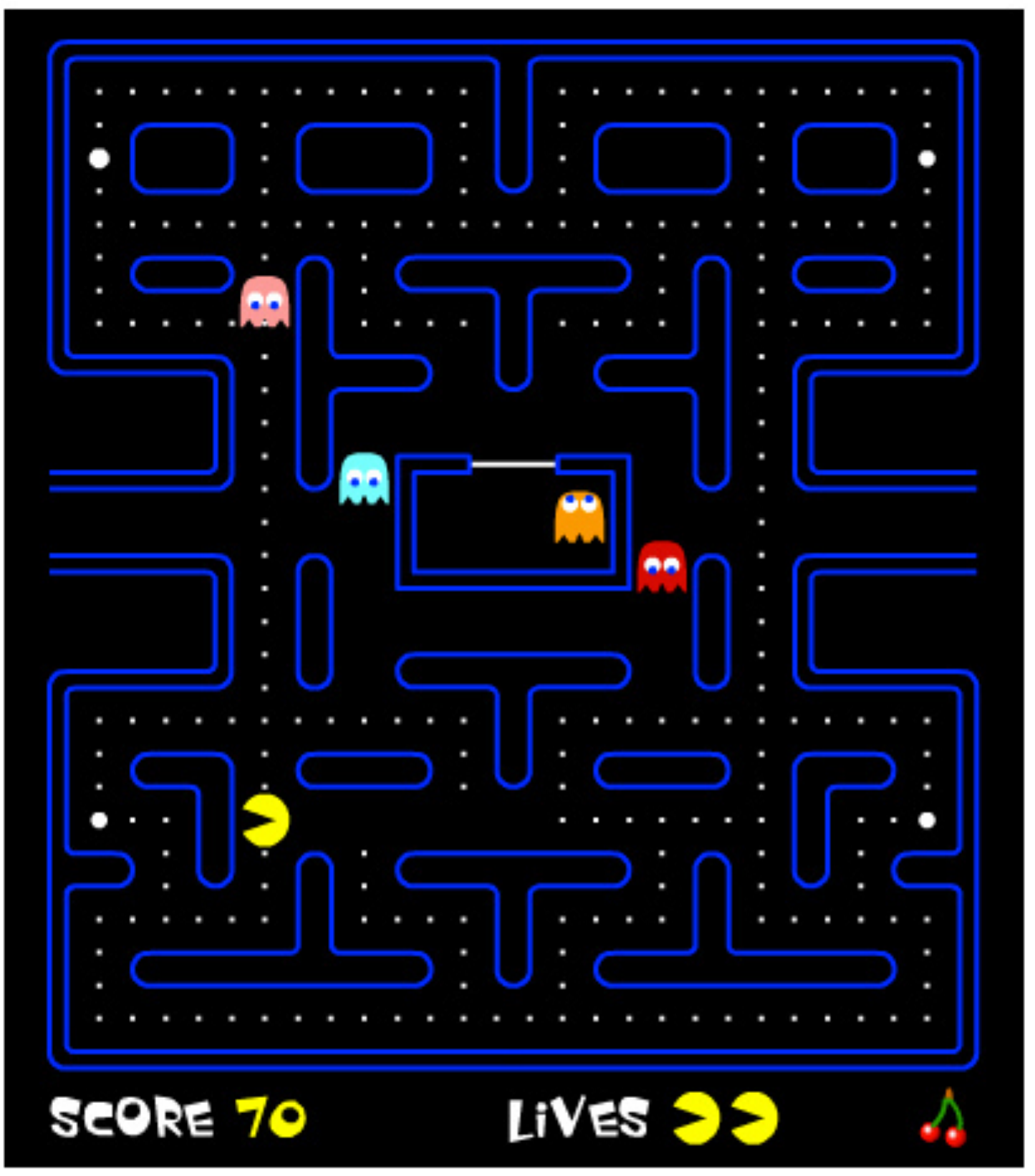}
\caption{The Pac-Man arcade game.}
\label{fig:pacman}
\end{figure}

\subsection{Pac-Man}
\label{pacman}
The experimental domain for the teaching methods presented in this article is the game of Pac-Man. Pac-Man is a famous 1980s arcade game in which the player navigates a maze like the one in Figure~\ref{fig:pacman}, trying to earn points by touching edible items and trying to avoid being caught by the four ghosts.
In our experiments, we use a JAVA implementation of the game provided by the Ms. Pac-Man vs. Ghosts League~\cite{rohlfshagen2011ms}, which conducts annual competitions.
Ghosts in our setting will chase the player 80\% of the time and choose actions randomly 20\%.

The player and all ghosts have four actions --- move up, down, left, and right --- but some actions are occasionally unavailable due to the restrictions in the maze. Four moves are required to travel between the small dots on the grid, which represent food pellets and are worth 10 points each.
The larger dots are power pellets, which are worth 50 points each. When the player gets the larger ones,  the ghosts become edible for a short time, during which they slow down and flee the player.
\rev{Eating a ghost is worth 200 points (which doubles every time for the duration of a single power pill). Then the ghost respawns in the lair at the center of the maze.}\AF{\#41}
The episode ends if any ghost catches Pac-Man, or after 2000 steps. 

This domain is discrete but has a very large state space.
There are 1293 distinct locations in the maze, and a complete state consists of the locations of Pac-Man, the ghosts, the food pellets, and the power pills, along with each ghost's previous move and whether or not it is edible.
The combinatorial explosion of possible states makes it essential to approach this domain through high-level feature construction and Q-function approximation.

In this article, we follow previous work~\cite{torrey13} that adopted a high-level feature set (high-asymptote feature set) comprised of action-specific features. When using action-specific features, a feature set is really a set of functions $\{f_1(s,a), f_2(s,a), ...\}$.
All actions share one Q-function, which associates a weight with each feature.
A Q-value is $Q(s,a) = w_0 + \sum_i w_i f_i(s,a)$.
To achieve gradient-descent convergence, it is important to have the extra bias weight $w_0$ and also to normalize the features to the range $[0,1]$.

For the state representation, we define a feature set which consists of $7$ features that count objects at a range of distances from Pac-Man maze, as we used (and defined) in previous work~\cite{torrey13}. 

A perfect score in an episode is 5600 points, but this is quite difficult to achieve (for both human and agent players).
An agent executing random actions earns an average of 250 points. The 7-feature set allows an agent to learn to catch some edible ghosts and achieve a per-episode average of 3800 points.

\section{The Teaching Task} 
\label{sec:theteachingtask}
In this section we attempt a more formal understanding of a teaching task that is based on action advice. The necessary notation is presented in Table \ref{tab:notation}.
\subsection{Definitions}
\label{sec:definitions}

\begin{table} 
    \centering
\scalebox{0.8}{
  \begin{threeparttable}
    \caption{Notation used in this article}
    \label{tab:notation}
\begin{tabular}{|c|c|c|c|}
\hline 
\backslashbox{Notion\tnote{\textdagger}}{Agent} & Student (Acting) & Teacher (Acting) & Teacher (Advising) \\ 
\hline
Index used & NONE & $\Sigma$  & $T$ \\ 
MDP & $M$& $M_\Sigma$\tnote{\textasteriskcentered}  & $M_T$  \\ 
Action Set &$A$  & $A_\Sigma$ & $A_T$  \\ 
State Space & $S$& $S_\Sigma$ & $S_T$  \\ 
Reward & $R$& $R_\Sigma$\tnote{\textbullet} & $R_T$  \\ 
Value Function & $Q$& $Q_\Sigma$ & $Q_T$  \\ 
Policy & $\pi$& $\pi_\Sigma$ & $\pi_T$  \\ 
Agent Goal & $L$& $L_\Sigma$\tnote{\textbullet} & $L_T$  \\ 
\hline 
    \end{tabular}
    \begin{tablenotes}
    \item[\textdagger] A teacher agent may not have a teaching value function $Q_T$, relying in a hand-coded or heuristic teaching policy
    \item[\textasteriskcentered] If the teacher has learned to act in the same MDP as the student, $M_\Sigma=M$
       \item[\textbullet] In this work we assume $R_\Sigma=R$ and $L_\Sigma=L$. All agents acting in the task have the same rewards and goals.  
    \end{tablenotes}
  \end{threeparttable}   }
  \end{table}

\extVersion{
\theoremstyle{definition}
\begin{definition}{\textbf{Actor}}
An actor agent (not to be related with the actor-critic RL architecture) is the agent \emph{acting} in the environment. Moreover the actor agent is also possibly learning from its interaction with the environment. The actor agent can be anything from a software agent to a robot or a human and its role is certainly one of the following:
\begin{itemize}
\item An agent without any teacher in the environment
\item A teacher agent while it is learning the task on its own (acting)
\item A student agent receiving advice from a teacher agent
\end{itemize} 
\end{definition}
}

\theoremstyle{definition}
\begin{definition}{\textbf{Student}}
A student agent is an agent acting in an environment and capable of accepting advice from another agent\footnote{In this work we assume that a student agent always follows the given advice}.
\end{definition}

\theoremstyle{definition}
\begin{definition}{\textbf{Teacher}}
\label{def:teacher}
\rev{A teacher agent is an agent capable to execute and inform a teaching policy (see Definition \ref{def:policy}) to provide action advice to a student agent acting in a specific task.}\AF{\#16}
\end{definition}

\theoremstyle{definition}
\begin{definition}{\textbf{Acting Task}}
The acting task is the task for which the teacher gives advice and can be defined as an MDP of the form $M = \langle S,A,T,R,\gamma \rangle $ on an environment $E$. 
\end{definition}

\theoremstyle{definition}
\begin{definition}{\textbf{Teaching Task}}
The teaching task is the task of providing action advice to a student agent to assist him in learning faster or learning better the acting task. Any teaching task is accompanied by a finite advice budget, $B$.
\end{definition}


\rev{
\theoremstyle{definition}
\begin{definition}{\textbf{Teaching Action Space}}
\label{def:teacheractions}
Given the action space $A$ of the acting task, the action space of the teacher in timestep $t$ is:

\[ A_T = \left\{ 
  \begin{array}{l l}
     \{a,\perp\} &, \quad \text{$b_t>0$}\\
    \{\perp\} &, \quad \text{$b_t \leq 0$}
  \end{array} \right.\]
where $a\in A$ an action of the acting task given as advice and the no advice action, $\perp$, meaning that the teacher will not give advice in this step allowing the student to act on its own. $b_t$ is the advice budget left in time-step $t$.
\end{definition}

\theoremstyle{definition}
\begin{definition}{\textbf{Teaching State Space}}
\label{def:statespace}
The teacher agent state space in timestep $t$ has the following form:
\begin{equation}
\label{statespace}
S_T = \langle\Theta_t,b_t,Q_\Sigma\rangle,
\end{equation}
where $\Theta$ is a tuple containing any knowledge we can have for the student and its MDP in timestep $t$. If the student's MDP is $M_\Sigma = \langle S,A,T,R,\gamma \rangle$ and the teacher observes the current state of the student, $s_t \in S$, reward, $r_t \in R$ and action $a_t \in A$ then $\Theta_t = \langle s_t,r_t,a_t \rangle$.
\end{definition}

\theoremstyle{definition}
\begin{definition}{\textbf{Teaching Policy}}
\label{def:policy}
A teaching policy, $\pi_{T}$, is a deterministic policy of the form: 
\begin{equation}
\label{policy}
\pi_{T}(S_T) \rightarrow A_T
\end{equation}\AF{\#43}\AF{\#18}
where $S_T$ and $A_T$ are the teaching state and action spaces respectively (see Definitions \ref{def:statespace} and \ref{def:teacheractions}). 
 \extVersion{e.g, one teacher agent may have $\Theta = \langle S,R \rangle$ which means that the teacher knows only the current state and reward of the student agent but it does not know the action it took, the transition function or the $\gamma$ parameter of the MDP.}
\end{definition}
}
The teaching policy actually transforms the acting policy, $\pi_\Sigma(s,a)$, of an actor agent (expressed through its respective state-action value function, $Q_{\Sigma}$), to a policy producing advice under budget. We should also note that a teaching policy will usually (\cite{torrey13,zimmer,zhan_online_2015}) set $a=\arg\max_a(Q_{\Sigma}(s,a))$ which means that the teaching policy is greedy with respect to the acting value function, $Q_\Sigma$.\AF{\#17} 

As a minimal example of the proposed formulation, the Importance Advising method~\cite{torrey13} which uses the state importance criterion (see Section \ref{background2}) can be said to \rev{use a teaching state space, $S_T = \langle\Theta=\{s\},B,Q_{\Sigma}\rangle$} \AF{\#18} as it requires knowledge only about the current state, $s$, of the student, the advice budget, $B$ and an acting policy, $Q_{\Sigma}$ from which it produces advice.

\subsection{Learning to Teach}
\label{sec:learningtoteach}
The definitions presented in subsection \ref{sec:definitions} apply to any teacher agent even if it advises based on a heuristic function. In the following, we focus on teachers that use RL to \emph{learn} a teaching policy (i.e., advice distribution policy).

In its most simplified version, the learning to teach task employs two agents: the teacher and the student. In the \emph{first learning phase} a teacher agent has the role of the actor: it learns the acting task alone. It observes a state space $S_\Sigma$ and has an action set $A_\Sigma$. Based on a reward signal $R_{\Sigma}$ received from the environment, it learns a policy $\pi_{\Sigma}$ to achieve the acting task goal $L_{\Sigma}$. In our context, this first learning phase can be seen as the \emph{advice production} phase since the teacher learns the policy that will be used to advise a student later on.

At any time-step $t$ the teacher agent may have to stop acting and a new agent, \emph{the student} enters the acting task and the corresponding environment.

Consequently, the teacher agent has to now \emph{learn and use a teaching policy} for the specific task to achieve the teaching goal, $L_T$. \rev{Additionally to the definitions given in Section \ref{sec:theteachingtask}}, this \emph{second learning phase} (learning a teaching policy), requires realizing and formulating the following:
\begin{itemize}
\remove{State Space. The state feature set of the teacher should include the state features of the student, allowing the direct use of the teacher's acting policy, $\pi_\Sigma$. This way advising will also depend on the current student state.}
\remove{Action Set. The action set of a teacher $A_T$ can have the two actions \{advice, not advice\} as in Definition \ref{def:teacheractions} also, taking into account \AF{\#26 is there something wrong here?} the remaining advice budget. \extVersion{When the teacher selects the advice action it makes a greedy selection over its acting value function, $Q_{\Sigma}(s,a)$ to produce the action advice.  When not advising it leaves the student act on its own policy. We should note here that although the teacher is internally greedy with respect to the acting goal, $L_{\Sigma}$, it has no restriction on being greedy with respect to its new teaching goal, $L_T$ i.e. learning its teaching policy will certainly require exploration of the teaching state space.} }
\item Return Horizon. Even if the teaching task is formulated as an episodic one, the teaching episode, also referred as a session, is not necessarily matching the student's learning episode. The teacher's episode scope is greater and could track several learning episodes of the student.
\item Reward signal. A different return horizon implies a different task goal and consecutively the teacher's reward signal can be different from the student's (e.g., encouraging more the learning progress of the student over its absolute learning performance). \extVersion{The special interesting case where $R_{\Sigma} = R_T$ could be valid too, however this articles focuses on task-independent teaching reward signals so that the proposed formulation can be applied to any task, with the less possible modifications.
\item Advice Budget. The teacher has to stop advising after giving all its $B$ advice while its only available action then is no advice.}
\end{itemize}

Moreover, defining the teacher's state space as a superset of the student's state space (see Definition \ref{def:statespace}) indicates one more difficulty of the learning to teach task. From the teacher's point of view the student can be considered a time-inhomogeneous Markov-Chain (MC)~\cite{stroock2004introduction}, $X=(X_t:t\geq0)$. This is because the transition matrix $P_t$ of the student's MC is dependent on time, since it is learning and constantly changing its policy over time. The time inhomogeneity of this MC poses significant difficulties in handling the problem theoretically. Homogenizing this MC by defining it as a space-time MC, $(X_t,t)$ can make practical solutions feasible but still theoretical treatment is difficult (e.g., no stationary distributions exist in this case).

\remove{Concerning the teacher's reward, a special interesting case where the acting and teaching reward signals are the same ($R_{\Sigma} = R_T$), could be valid too. However, $R_{\Sigma}$ is task-dependent and this article focuses on task-independent teaching reward signals so that the proposed formulation can be applied to any task.}

In general, every \emph{learning task} can have its corresponding \emph{teaching task} which could be thought as its dual. As learning to act in a specific task and teaching that task can be considered different tasks, they have their own goals and consequently, are ``described'' by different reward signals. 

As an example, in~\cite{zimmer} a teacher agent for the mountain car domain has a different reward signal to that of the student, encouraging teaching policies that help the student reach its goal sooner. 

Learning a teaching policy, as this is described above, could be modelled by many different types of Markov Processes. 
However, none of the classic MDP formulations 
completely models the specific learning problem as a whole either by not handling the non-stationarity of the problem or by not handling the specific budget constraint imposed on the advising action. This fact is the main motivation of Section \ref{sec:distributeadvice}, where we present our proposed method for learning teaching policies.

\section{Learning to Produce Advice} 
\label{sec:produce}
In this section, we focus on the advice itself and its production (not its distribution). The main challenge in producing advice based on the Q-Values of an RL value function is that these values are valid only if the policy they represent is fully followed, not when this policy is sparingly sampled to produce advice. 

Based on previous methods in the literature (see Section \ref{sec:related}) the most common teacher's criterion for selecting which action to advise is $\pi_{\Sigma}(s)=argmax_a{Q_\Sigma(s,a)}$, that is, greedy selection of the best action based on the teacher's acting value function. However, the value of $Q_\Sigma(s,a)$ is not correct under the advising scenario since it is accurate only if the student \textit{will continue following teacher's acting policy, $\pi_\Sigma$ thereafter}. Unfortunately, this is usually not the case in our context since the student, after receiving advice, will often continue for a long period using its own policy exclusively. Even worse, in the early training phases---when advice is needed the most---the student's policy will be vastly different from the teacher's.
 

This realization is even more important if we consider how different the teacher and the student agents are allowed to be in our context. Consider a human student receiving advice in the game of Pac-Man. Human players often play fast-paced action games in a myopic and reactive manner, seeking short-term survival and not a long-term strategic advantage. 

In that case, a human student infrequently advised by a policy learned using a high $\gamma$ value close to 1 will often be mislead to locally sub-optimal actions because these actions may be highly valued for the teacher's far-sighted policy. The human player will probably not follow such a policy thereafter and he has therefore been misled to an action that would be useful only if he would also follow the rest of the teacher's acting policy too. 
 
Ideally, we would like to use a teacher's acting policy that would be mostly \emph{invariant} to the student's particularities. Such a teacher's policy would advise actions that are good on average, whatever policy is followed thereafter by the student and whatever its internals and parameters are (e.g. $\gamma$ value) etc. 
 
In this article, we propose that the above considerations should affect the way we learn policies intended for teachers. Selecting a specific policy for advising, the RL algorithm producing it and its parameters, form a \textit{model selection problem} for RL teachers.

\subsection{Model Selection for Teachers}
\label{sec:modelselection}
In this section, we want to investigate how factors such as the teacher's $\gamma$ value (see Section \ref{sec:rl}) influence advice quality for students that can possibly have very different characteristics (e.g, a myopic student and far-sighted teacher). This is important in order to understand which teacher-agent differences affect the teaching performance the most.

To assess the influence of the $\gamma$ value in the teaching process, we experiment using an RL algorithm like R-Learning~\cite{rlearning,sutton98}, which does not use a $\gamma$ value for the calculation of state-action values and relies on estimating the average reward received by the student, using its policy from any state and thereafter.

Specifically, R-Learning is an infinite-horizon RL algorithm where a different optimality criterion is used such that the value $Q(s,a)$ given action $a$ and state $s$ under policy $\pi$ is defined as the expectation:
\begin{equation}
\label{criterion}
Q(s,a) = \sum_{k=1}^{\infty}{E_\pi\{r_{t+k}-\rho^\pi \} }
\end{equation}
Where $\rho^\pi$ is the \emph{average} expected reward per time step under policy $\pi$. The intuition behind R-Learning is that in the long run the average reward obtained by a specific policy is the same, but some state-action pairs receive better-than-average rewards for a while, while others may receive worse-than-average rewards. This transient, the difference to the average reward received, $\rho^\pi$, is what defines the state-action value. \rev{To keep a running estimate of the average reward, R-Learning uses a second update rule, and one more parameter, $\beta$, for the learning rate of that update.} \AF{\#22} 


Using R-Learning to learn a teacher's acting policy along with the rest of the experiments presented in Section \ref{sec:expproduceadvice}, we can assess the importance of $\gamma$ value and $\gamma$ value mismatch between student and teacher. Moreover, we assess other factors that possibly influence the quality of advice such as the performance of the teacher in the acting task, its performance variance and a possible relation of its average \emph{td-error} \cite{sutton98}, with the quality of advising.

As defined in \cite{sutton98} the td-error, $\delta_t$, represents the value estimation error of a value function for a specific state $s$ and action $a$ in time $t$. For the Q-Learning \cite{watkins92} algorithm that is:
\begin{equation}
\label{td-error}
\delta_t = r + \gamma\max_{a}Q(s_{t+1},a)-Q(s_t,a_t).
\end{equation}

\noindent This is also part \AFold{didn’t get the word you suggested} of the Q-Learning update rule. Furthermore, by dividing (\ref{td-error}) with the previous value estimation, $Q(s,a)$, we get the percentage of error in relation with it, which we can call td-error percentage:
\begin{equation}
\label{td-error-pct}
\delta^\%_t=\delta_t/Q(s_t,a_t),
\end{equation}

\noindent where $Q(s_t,a_t) \neq 0$. In our context, when the teacher uses an acting policy to produce advice it can still compute, for each student's experience, its own td-error just as it would do if it was actually making a learning update. In the same context, we can intuitively say that $\delta^\%_t$ represents the \emph{teacher's surprise}\footnote{\rev{Note that this definition of surprise, although similar, is different to that presented in \cite{white2014surprise} which normalizes for different learners and not for different state-action pairs.}\AF{\#24}} on its new estimation of a state-action value. 

Consequently, a teacher with high average td-error percentage, $\overline{\delta^\%}$, is a teacher with more unreliable value estimation, and therefore, it can be less suitable for a teacher since its action suggestion is based on a non-converged value function. 

\begin{table*}[t]
\caption{\rev{Average Score, Coefficient of Variation (CV) and Standard Deviation of $\gamma$-specific Q-Learning agents and R-Learning, acting alone in 500 episodes, ordered by score.}\AF{\#9 \#2}} 
\centering
\scalebox{0.85}{
\begin{tabular}{|l|l|c|c|c|}
 \hline
&\textbf{Gamma}&	\textbf{Average Episode Score} $\blacktriangledown$ &\textbf{Coefficient of Variation}&\textbf{\rev{Std. Deviation (SD)}} \\ \hline
Q-Learning &  &	&	&\\
& 0.9&	3633.78&0.33&1189.89\\
&0.05 & 2754.48	 &0.44&1132.77\\
&0.2 &2668.36&0.47&1254.13\\
& 0.999 &	2608.04&0.28&730.25\\
&0.6 & 2585.26	& 0.48&1240.92\\
R-Learning & & 	& &\\
&-  & 2493.17&0.28&698.09	\\
 \hline
\end{tabular}
}
\label{teacherperf}
\end{table*} 

\begin{figure*}[t]
\centering
\hspace{-2.4em}\includegraphics[width=0.75\textwidth]{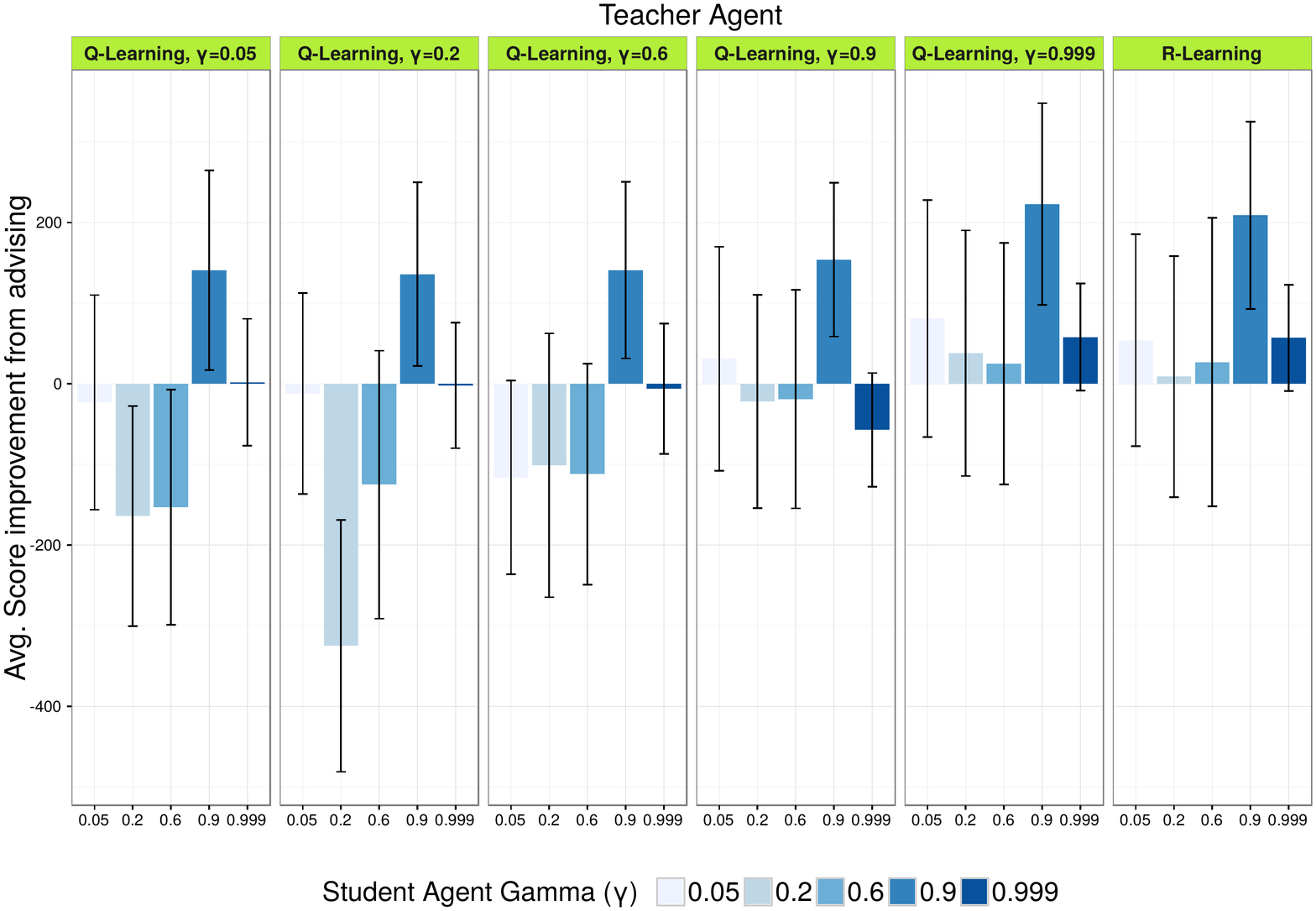}
\caption[Caption for LOF]{Average score improvement from 30 trials of each teacher-student pair compared to no-transfer (no advising). Negative bars indicate negative transfer (average score decrease). Error bars indicate 95\% confidence intervals (CI) of the means. Non-overlapping CIs indicate statistically significant differences of the means whereas overlapping CIs are inconclusive\footnotemark}

\label{fig:RLearning}
\end{figure*}

\subsection{Experiments and Results}
\label{sec:expproduceadvice}
Based on the discussion in the previous section (Section \ref{sec:modelselection}) the main goal of the following experiments is to find the teacher's policy parameters (such as $\gamma$) that affect the quality of advice most for different student parameters. The experimental design is as follows. In the first phase, we created $\gamma$-specific teachers by training five Q-Learning agents and one R-Learning agent for 1000 episodes. The Q-Learning agents had all the same parameters, except $\gamma$, which took values in $\{0.05,0.2,0.6,0.9,1.0\}$. The rest of their parameters were the same and fixed, specifically $\epsilon=0.05$ and $\alpha=0.001$ (same with previous work~\cite{torrey13})\AFold{(same with.. ok?}. The $\lambda$ parameter accounting for eligibility traces was set to zero so that the effect of experimentally controlling the $\gamma$ parameter is isolated. Finally, the parameter $\beta$ of R-Learning was set to 0.0001 (preliminary results found it produced good results in Pac-Man). 

After training for 1000 episodes, the $\gamma$-specific Q-Learning teachers and the R-Learning teacher were evaluated on 500 episodes of acting alone in the environment. \rev{We calculated their average episode score and the coefficient of variation of these scores as both being possible determining factors of advice quality. }\AF{\#37} Coefficient of variation was used as a measure of score discrepancy as it shows the extent of variability in relation to the mean of the score, allowing a more clear comparison of variance between methods with different average performance. It is a unit-less measure calculated as $c_v = \frac{\sigma}{\mu}$.

In Table \ref{teacherperf} we can see their average episode score on 500 episodes along with the coefficient of variation of that score. R-Learning had significantly worse average \emph{acting} performance than all versions of Q-Learning. Interestingly, episodic Q-Learning (with $\gamma$ close to 1) did not perform as well as expected. Moreover, a very low $\gamma$ value (0.05) came up second, showing that a myopic RL agent can perform well in Pac-Man. This result indicates the highly stochastic nature of the game where reactive short-sighted strategies, based more on survival, can perform better than far-sighted strategies. 

After the initial training and the evaluation of the acting policies they learned, these agents could be used as teachers for tabula-rasa student agents. In these experiments we used a simple fixed teaching-advising policy called Every-4-Steps for all teachers since we focus only on the quality of the advice itself and not on the quality of its distribution to the student (teaching policy). 

In the Every-4-Steps teaching policy, the teacher gives one piece of advice to the student every four steps. Using this fixed advising policy we can test and compare the efficacy of the advice when this is not given consecutively, thus testing how useful the advice is when the student does not take a complete policy trajectory from the teacher, but has to use its own policy in between.

Using the teaching policy Every-4-Steps and a budget of $B = 1000$ advice we ran 30 trials of advising learning students for \emph{each} $\gamma$-specific teacher-student pair. Specifically, the $\gamma$ parameters of these teacher-student pairs come from the Cartesian product $\{0.05,0.2,0.6,0.9,0.999,-\} \times \{0.05,0.2,0.6,0.9,0.999\}$ (30 pairs), where the R-Learning teacher in the first set is denoted with a \quot{-} since it does not have a $\gamma$ value. 

\footnotetext{For the non-conclusiveness of overlapping confidence intervals, a simple and intuitive explanation can be found in \url{https://www.cscu.cornell.edu/news/statnews/stnews73.pdf}}In Figure \ref{fig:RLearning} we can see the \rev{average} \AF{\#12} performance of each teacher-student pair compared to the same student not receiving advice at all. Combining these results with Table \ref{teacherperf} of the teachers' performance when they were acting alone, we can see that \emph{the best performer is not the best teacher}, with \emph{best} defined as the best average score when acting alone in the task. The best example of this is R-Learning whose average score was the worst than any $\gamma$-specific Q-Learning agent, however, as we can see in Figure \ref{fig:RLearning} is almost as good of a teacher as the $\gamma=0.999$ Q-Learning teacher. R-Learning advising improved all student's score whatever their $\gamma$ value, while not resulting in a negative transfer for any of them. 
 
Moreover, we can see a pattern where the lower the coefficient of variation (CV) for the acting performance is, the better the teacher, indicating that CV can be an important 
criteria in model selection for teachers. This is non-trivial since average agent performance (and not its variance) is the dominant model selection criteria adopted in most of the relevant literature in RL. Performance variance expressed by CV seems especially important in our context, that of sparse advising, where the advice should be good whatever the next actions of the student will be.
 
Based on the results presented here, we can not \AF{\#26 what is wrong with.. can not?} observe any particular pattern relating teaching performance with the $\gamma$ values of a teacher-student-pair. Interestingly though a $\gamma=0.999$ teacher is not the most helpful for a $\gamma=0.999$ student. Even more, a $\gamma=0.2$ for a $\gamma=0.2$ student results to significant negative transfer. The teacher with the episodic $\gamma$ value, $\gamma = 0.999$ and the no discounting R-Learning one were the most helpful to all students showing that R-Learning can perform well in settings where the student's $\gamma$ is unknown or varying, such as in the case of human students.

Having identified the possible use of R-Learning for producing acting policies suitable for advising and the importance of performance CV to model selection, we conducted one more experiment between identical teachers. 

Specifically, we independently trained 30 Q-Learning teachers with the same parameters, feature sets and characteristics for 1000 episodes. Due to their different experiences and the stochasticity of the game they naturally learned different policies (i.e., final feature weights in their function approximators). Then, the trained teachers played alone for 500 episodes and we recorded their average performance, average performance variance as also their average TD-error percentage, $\Delta^{pct.}_t$, as this was defined in Section \ref{sec:modelselection}. We then used the Every-4-step teaching policy with each one of them advising a standard Sarsa \cite{Rummery94} student who would learn the task for 1000 episodes. Finally, we recorded the student's average score.

\begin{figure}[t]
\centering
\includegraphics[width=0.35\textwidth]{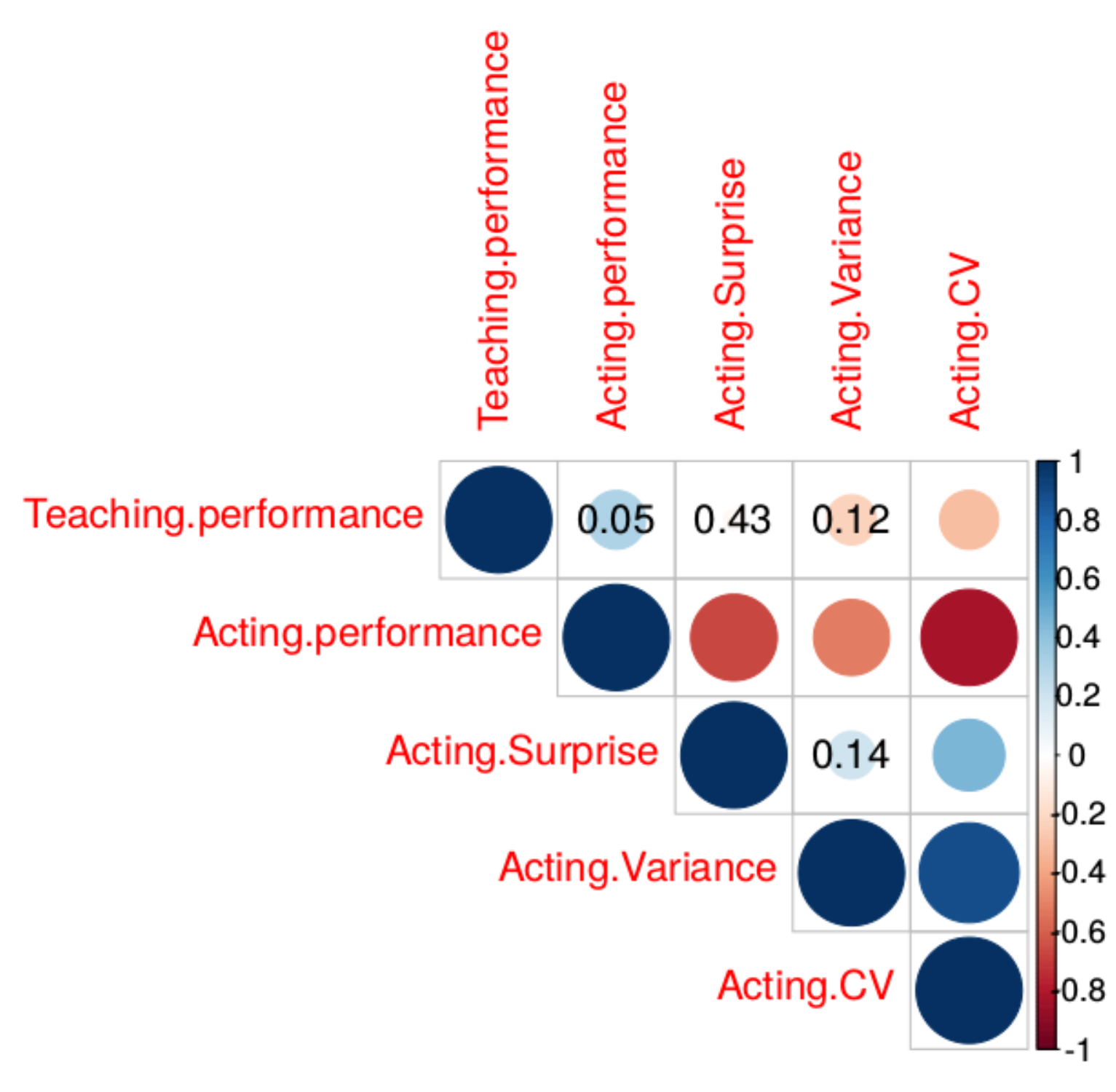}
\caption{Correlation plot of critical factors influencing teaching performance for 30 Q-Learning teachers. The circle size indicates correlation strength and the color its sign (direction). Statistically significant correlations are at p<0.05. Insignificant correlations are marked with their $p$ value.
}
\label{fig:RLearningCor}
\end{figure}

In Figure \ref{fig:RLearningCor} we can see a correlation plot of the factors mentioned above using a one-tailed non-parametric Spearman correlation test at $p<0.05$. 
\extVersion{Size and color of the circles indicate strength and direction of the correlation respectively. The figure also reports the p-value for the stat. insignificant correlations. }
Confirming the previous results we can see the negative and statistically significant relation of CV to teaching performance with $r=-0.3$. Acting performance also has a medium and positive correlation of $r=0.3$ with teaching performance (student's score) but it is \rev{statistically} insignificant on the limit. 
By weighing average performance in its calculation, CV has a stronger relation to teaching performance than standard statistic variance. Moreover, we see that teacher's surprise, $\Delta^{pct.}_t$ relates strongly ($r=-0.66$) and negatively to the acting performance of the teacher and not to its teaching performance ($r<-0.05$).

\section{Learning to Distribute Advice} 
\label{sec:distributeadvice}
In this section we change focus from advice production to advice distribution, learning a teaching policy in order to most effectively distribute the advice budget.

\subsection{Constrained Exploitation Reinforcement Learning}
We attempt a more natural formulation of the AuB learning problem described in Section \ref{sec:learningtoteach} by 
identifying it as an instance of a more generic reinforcement learning problem. This RL problem can be simply described as learning control with constraints imposed on the \emph{exploitation} ability of the learning agent. These constraints can either be a finite number of times the agent can exploit using its policy, possibly states where it is only allowed to explore, or even perhaps a task where it is costly to have access to an optimal policy and we are allowed to use it only for a limited number of times.  
How does this RL problem relates to the learning to teach problem? The first insight is that the advise/no-advise decision problem has a striking resemblance to the core exploration-exploitation problem of RL agents. Consider the learning to teach problem. We can view the problem as follows: When the teacher agent is advising it is actually acting on the environment, that is because an obedient student agent will always apply its advice thus becoming a deterministic actuator for the teacher. In the case of a non-obedient student, the teacher could be said using a stochastic actuator. 

Consequently, we can view the teacher agent as an \emph{acting} agent using a student agent as its actuator for the environment. Moreover, the teacher is acting greedily by advising its best action; thus, it exploits. Under this perspective, with advice seen as action, how could we view the no advice action of a teacher? The no advice action can be seen as \quot{trusting} the student to control the environment autonomously. Thus, choosing not to advise in a specific state can be seen as denoting that state to be non-critical with respect to the remaining advice budget and the student's learning progress, or denoting a lack of teacher's knowledge for that state. From the teacher's point of view, not advising can be seen as an exploration action. So controlling when \emph{not to advise} can be seen as a directed exploration problem in MDPs. Imposing a budget constraint, that is a constraint on the number of times a teacher agent can advise (i.e., exploit) is a problem of \emph{constrained and directed exploitation}.

We will consider a simple and motivating example of such a domain. In a grid world $10 \times 10$ a robot learns an optimal path towards a rewarding goal state while it should keep away from a specific damaging state. The robot is semi-autonomous, it can either control itself using its own policy or it can be teleoperated for a specific limited number of times. For the robot's operator, what is an optimal use of this finite number of control interventions? What are the states that it would be best to control the robot directly, leaving control of the rest to the robot?

Similarly to the previous example, learning and executing advising policies in a game can be another example of the constrained exploitation problem, which is also the main focus of this article. For example, in a video game like Pac-Man, a game hints system plays the role of the external optimal controller with a limited intervention budget. Such a hint system could suggest actions to human players---when these are most necessary---depending also on the player's policy.\METold{Example in Pac-Man?}\AFold{I added this one but I didn’t remove the previous example because I think it is a pity to not show that advising is just an instance of this class of problems, and that more generic control problems can fall into the same category (CERL). Do you agree?}

In the rest of this section, we use the term exploitation where one can think of advising and the term exploration when not-advising, focusing on the broader learning problem.

\subsection{Learning Constrained Exploitation policies}

Formulating the constrained exploitation task as a reinforcement learning problem itself first requires defining a horizon for the returns. This horizon should be different from that of the actual underlying task (e.g., Pac-Man) because a) if the underlying task is episodic then the scope of an exploration-exploitation policy is naturally greater than that and spans across many episodes of the learning agent b) if the underlying task is continuing or requires several training episodes for the student, the exploration-exploitation policy may have to be evaluated in a shorter (finite) horizon (e.g., for the first $x$ training episodes).  
The importance of exploration is usually limited in the late episode(s) where the student may have already converged to a policy. A teaching policy should be primarily evaluated for a training period where advice still matters. 

Concerning the return horizon of a constrained exploitation task (and similarly to~\cite{zhan_online_2015} but in a different perspective), we propose algorithmic convergence~\cite{zhan_online_2015} as a suitable \emph{stopping criterion} for an exploration-exploitation policy. This defines a meaningful horizon for exploration-exploitation tasks since their goal is completed exactly then, not in the end of an episode and not in the continuous execution of an RL algorithm---after convergence---where exploration may not affect the underlying policy anymore. We proceed by defining the Convergence Horizon Return.

\theoremstyle{definition}
\begin{definition}{\textbf{Convergence Horizon Return}}
\label{converge}
Let $G$ be the return of the rewards $r_t$ received by an exploration-exploitation policy, $Q$ the value function of the \textit{underlying} MDP and $\epsilon \in \mathbb{R}$ a small constant then:   
\begin{equation}
\label{ex_l}
G = \sum_{t = 0}^{T}{r_t}
\end{equation} 
where for the time step $T$ applies:
\begin{equation}
\label{ex_l2}
\left\vert{Q_{T+1}-Q_{T}}\right\vert - \epsilon \leq 0
\end{equation} 
\end{definition}

Given a small constant $\epsilon$ and the algorithmic convergence of the RL algorithm learning in the \emph{underlying} MDP, the quantity $\Delta Q = \left\vert{Q_{t+1}-Q_{t}}\right\vert - \epsilon \xrightarrow[t \to \infty]{} 0 $. The algorithmic convergence will be realized either if the learning rate $\alpha$ is discounted or if some temporal difference $\Delta_t$ of the underlying algorithm tends to $\epsilon$. 

Using the convergence horizon for the return of a teaching task too, the next question can be what are the rewards $r_t$ constituting the return of a teaching task. 

One possible goal for any teacher advising with a finite amount of advice would be to help minimize \emph{student's regret} with respect to the reward obtained by an optimal policy. However, since we do not assume such knowledge, and because there is a finite amount of advice, a better goal could be to advise based on the state-action \emph{value} of the advised action and not its immediate \emph{reward}. If the student was able to follow the rest of the teacher's policy after receiving advice, then the action $a=argmax_a{(Q_\Sigma(s,a))}$ for the current state $s$ would be the best possible. Consequently, we define the notion of \emph{value regret}.

\theoremstyle{definition}
\begin{definition}{\textbf{Value Regret}}
\label{regret}
In a convergence horizon $T$, the value regret, $R^V$ of an exploration-exploitation policy \rev{(i.e., teaching policy) with respect to both an acting policy $\pi^*$ obtained after the $T$ period and an acting policy (i.e., student's policy), $\pi^t$, in time step $t$ is:}
\begin{equation}
R^V = \sum_{t \in T}^{}[ max_{a}Q^*(s_t,a) - Q^*(s_t,\pi^t(s_t))],
\end{equation} 
\end{definition}

\noindent\rev{where $Q^*$ denotes the corresponding value function of $\pi^*$.}\AF{\#20} 

The intuition behind this definition of regret in our context (where the acting agent is the student) is that the best teacher for any specific student would ideally be the \emph{student himself}, when it would have reached convergence or its near-optimal policy. 

The important thing to note here is that because a student agent receives a finite amount of advice it cannot improve its asymptotic performance~\cite{zhan_online_2015}, consequently the evaluation of a teaching policy should ideally be based on the student's optimal policy and not to that of some probably very different teacher, because that is its \emph{sustainable optimality}. 

For example, consider two states in a teacher's acting MDP, $A$ and $B$. A student agent learning with a very simplistic state representation may observe these states as just one, $C$, and not differentiate between them. Then, the student's optimal action in state $C$ will have a different expected return than that obtained by the teacher from either $A$ or $B$. Its sustainable optimality is defined as to what is optimal given its simplistic internal representation. Any advice based on a finer representation may not be supported with consistency by the student in the long run. A teaching policy should be ideally evaluated on how much it speed up the student converging to its own optimal policy.\AFold{OK? about sustainable optimality}

In the next section we propose a reward signal for teachers based on Value Regret.


\subsection{The Q-Teaching algorithm}

\alglanguage{pseudocode}
\begin{algorithm*}[!t]
 
\caption{Q-Teaching}
{
\begin{algorithmic}[1]
\State Initialize $Q_T(s,a)$ arbitrarily \Comment teaching value function
\State Use existing $Q_\Sigma(s,a)$ or initialize it \Comment acting value function
\Repeat~(for each teaching episode) \Comment teacher-student session
	\State Initialize s
	\Repeat~(for each step)
	
		\State $a^* \gets \max_{a}Q_\Sigma(s,a)$
		\If {(Off-Student's policy Q-Teaching)} 
		\State $\hat{a} \gets \min_{a}Q_\Sigma(s,a)$ 
		\Else
		\State $\hat{a} \gets a$ \Comment where $a$ is the action announced by the student
		\EndIf
		\State Choose $a_T$ from $s_T$ using policy derived from $Q_T$ (e.g. $\epsilon$-greedy)
		\If {$a_T$ = \{advice\}} 
		\State Advice the student with the action $a^*$
		\ElsIf {$a_T$ = \{no\_advice\}} 
		\State Send a $\perp$ (no advice message) to the student 
		\EndIf
		\State Observe student's actual action $a$ and its new state and reward, $s', r$ 
		\State $Q_\Sigma(s,a)\gets Q_\Sigma(s,a)+\alpha[r + \gamma\max_{a'}Q_\Sigma(s',a')-Q_\Sigma(s,a)]$ \Comment possibly continue learning an acting policy
		\If {$a_T$ = \{advice\}} 
			\State $r_T \gets Q_\Sigma(s,a^*) - Q_\Sigma(s,\hat{a})$
		\ElsIf {$a_T$ = \{no\_advice\}} 
		\State $r_T \gets 0$
		\EndIf

		\State $Q_T(s,a_T)\leftarrow Q_T(s,a_T)+\alpha[r_T + \gamma\max_{a'}Q_T(s',a')-Q_T(s,a_T)]$
		\State $s\gets s'$
\Until{advice budget finishes OR reached the estimated convergence horizon episode of the student}
\Until{end of teaching episodes}
\end{algorithmic}}
   \label{alg:off}
\end{algorithm*}

The Q-Teaching algorithm described and proposed in this section is an RL advising (teaching) algorithm learning a teaching policy. For this, we propose a novel reward scheme for the teacher based on the value regret (see Definition \ref{regret}). 

The key insight of the method is that of rewarding a teaching policy with quantities of the form 
$max_{a}Q^*(s_t,a) - Q^*(s_t,\pi^t(s_t))$ where $\pi^t(s_t)$ is an estimation of the student's action in $s_t$ and $max_{a}Q^*(s_t,a)$ is the teacher's greedy action in $s_t$ (i.e., the action used for advice). This reward has a high value when the value of the greedy action is significantly higher than the value of the action that the student would take. This means that the teacher is encouraged to advise when the advised action is significantly better than the action the student would take. 

For terms of efficiency and to emphasize the value impact of the advising action, Q-Teaching 
rewards all no-advice actions with zero. The advantages of such a scheme is that the teacher's cumulative reward is based only on the value gain produced when advising and a teaching episode can finish when the budget finishes, not having to observe all the student's episodes after its budget finishes. From preliminary experiments, rewarding no advice actions too (which occur significantly more than the maximum $B$ advice actions) was overpowering the advice actions, resulting in an imbalanced expression of the two actions in the teaching value function.

Still, when advising, the teacher should estimate $Q^*(s_t,\pi^t(s_t))$ in order to compute its reward. The simplest solution is that since we do not have access to the value function of the student or its internals, we use the acting value function $Q_\Sigma$ of the teacher as an approximation for the optimal value function of the student, $Q^*$. To estimate $\pi^t(s_t)$ the teacher has several options. 
If the teacher is notified of the intended action of the student beforehand, it can use that to compute the reward. 
If we assume no knowledge of the student's intended action then some other estimation method for the student's intended action should be used. 
An example of such an estimation method is used in the Predictive Advice method~\cite{torrey13}.

While predicting the actual student's action ($\pi^t(s_t)$) is possible, there are other---simpler---choices for this estimation too. For example, the Importance Advising (see Section \ref{background2}) uses a very similar quantity for the advising threshold, of the form $max_{a}Q^*(s_t,a) - min_{a}Q^*(s_t,a)$. For Importance Advising, we can say that $\pi^t(s_t) = min{_a}Q^*(s_t,a)$---it pessimistically assumes the student will take the worst action, representing the risk of the state. The advantage of such an assignment is that it is based on a well-tested criterion~\cite{torrey13} and that it does not need knowledge of the student's intended action (desirable for most realistic settings). The disadvantage is that we have a less detailed reward which is also not adapting to the student's specific necessities but mostly, to the domain's characteristics.

Based on this dichotomy, we propose two versions of Q-Teaching (see Algorithm \ref{alg:off}), the off-student's policy Q-Teaching and the on-student's policy Q-Teaching. 
The on-student's policy Q-Teaching uses the value of the actual student’s action to compute the reward (thus it is directly influenced by its policy). We can intuitively say that on-student's policy Q-Teaching will advise when the student is mostly expected to act sub-optimally with respect to the acting value function of the teacher, $Q_\Sigma$. On the other hand, the off-student's policy Q-Teaching uses the criterion discussed above and the teaching policy is not \emph{directly} influenced by the policy of the student. Specifically, it is rewarding its teaching policy, $\pi_T$, at time-step $t+1$ with the q-value difference of the best action $a^*$ to the worst action, as these were found at time $t$.

\begin{figure*}[t]
	 \caption{Average student score in 1000 training episodes with teachers \textit{not knowing} the student's intended actions. The curves are averaged over 30 trials and the legend is ordered by total reward. The error bars represent the 95\% confidence intervals (CI) of the means. Non-overlapping CIs indicate statistically significant differences of the means whereas overlapping CIs are inconclusive}
	\begin{subfigure}{.5\linewidth}
	\centering
	\caption{Low-asymptote Sarsa student}
	\includegraphics[width=0.95\textwidth]{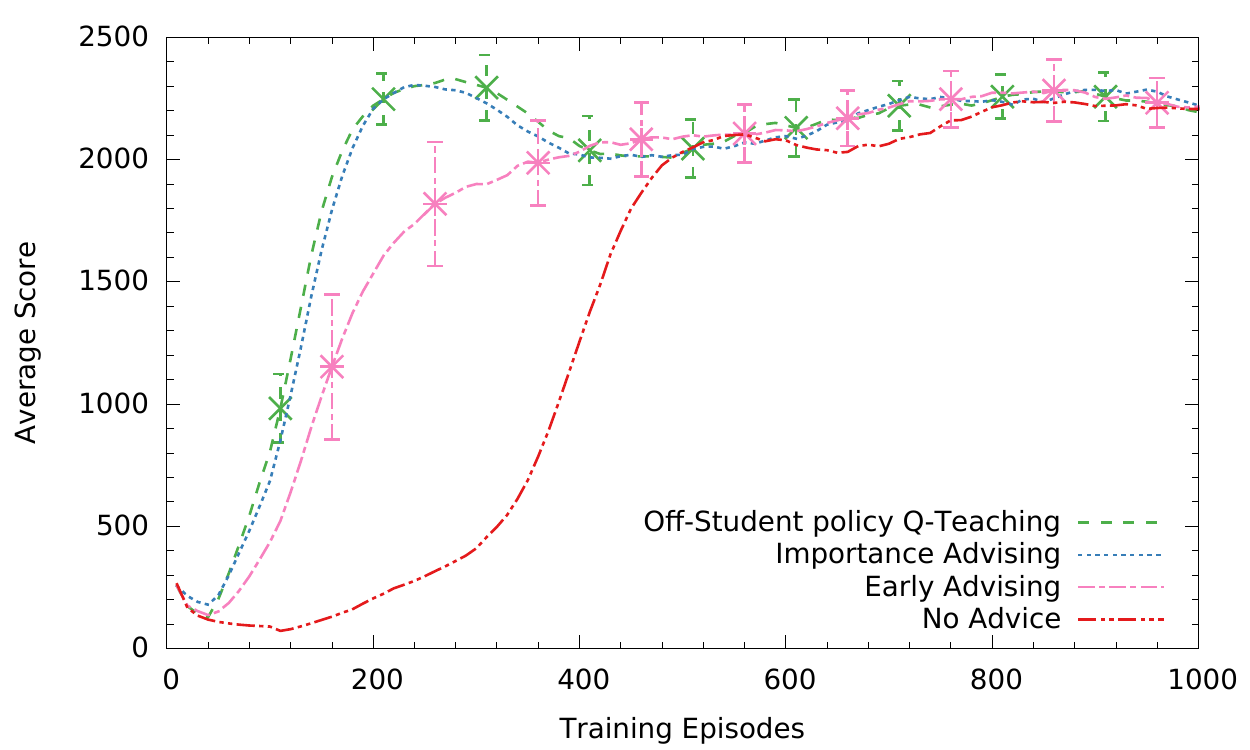}
	\label{fig:DepthS_student_agnostic}
	\end{subfigure}%
	\begin{subfigure}{.5\linewidth}
	\centering
	\caption{High-asymptote Sarsa student}
	\includegraphics[width=0.95\textwidth]{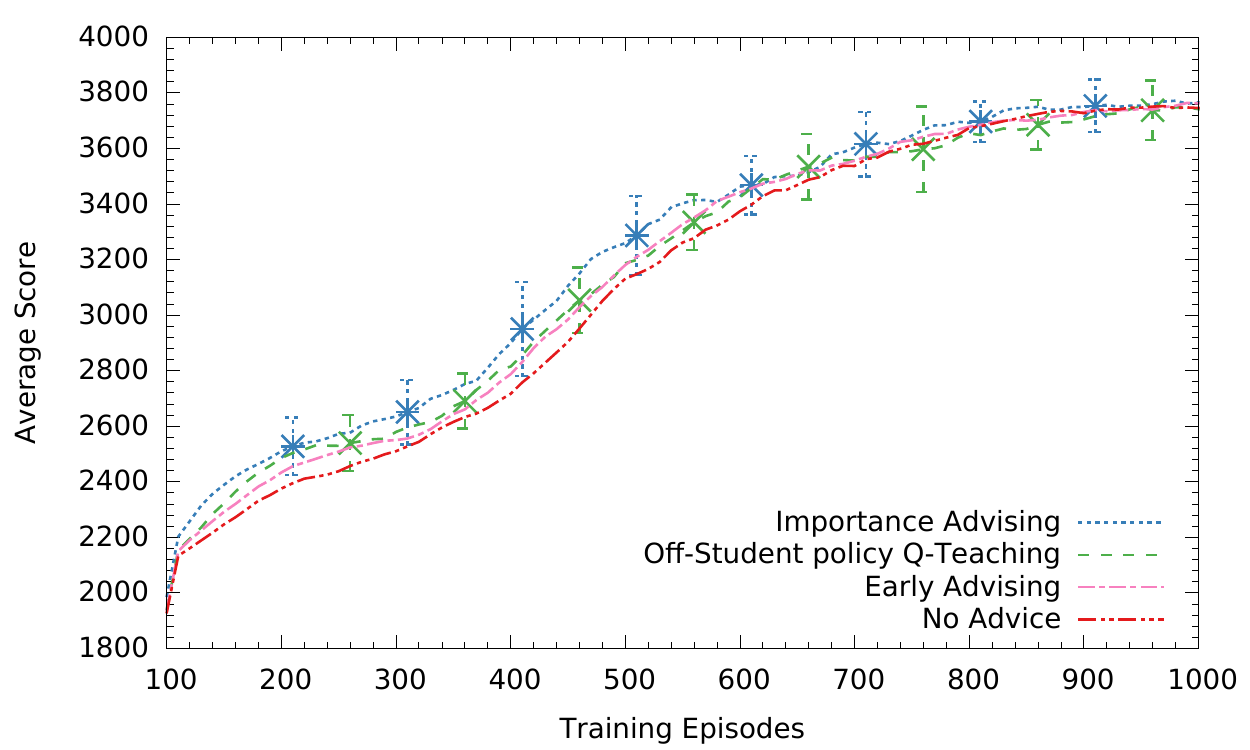}
	\label{fig:CustomS_student_agnostic}
	\end{subfigure}
	\end{figure*}

The Q-Teaching algorithm proceeds as follows (see Algorithm \ref{alg:off}). A teacher agent enters an RL acting task to learn an acting policy. It initializes two action-value functions, $Q_\Sigma$ and $Q_T$, the acting value function and the teaching value function respectively (lines 1-2). Of course, it can also use an existing acting value function.

Being in time step $t$ and state $s$ the teacher queries its acting value function for the greedy action in that state (line 6).
Depending on whether we use the off-student's policy or the on-student's policy Q-Teaching, the teacher sets a baseline action, $\hat{a}$, to either the worst possible action for that state or to the action just executed by the student (lines 8-12).

Then, the teacher chooses an action from $A_T = \{advice, not\_advice\}$ based on $Q_T$ and its exploration strategy.
If the teacher chooses to advise (line 13) it gives the action $a^*$ as an advice to the student agent. \rev{If the teacher chooses not to advise, the student will proceed with its own policy.}\AF{\#44} 

In line 19 the teacher observes the student's actual action $a$ and its new state and reward, $s', r$. Once again, the student may be the teacher himself, in this case, it observes its own action which was taken based on $Q_\Sigma$ and its exploration strategy.

In line 20, the first Q-Learning update takes place for the acting value function $Q_\Sigma$ based on the environment's reward. For the teaching value function update, the teacher's reward, $r_T$ is calculated first, based on the \emph{freshly updated values} of the best and baseline actions, $a^*$ and $\hat{a}$ respectively (lines 21-25).

Finally, a Q-Learning update for the teaching value function takes place based on the reward $r_T$ and the algorithm continues in the same way until whatever of the following two events comes first: Either the advice budget finishes or the student reaches a learning episode which we have predetermined as its convergence horizon. These complete one learning episode or session for the teacher.


In this version, the Q-Teaching algorithm is based on the Q-Learning algorithm, although in principle any RL algorithm could be used for the underlying learning updates of Q-Teaching. However, if an off-policy RL algorithm such as Q-Learning is chosen for the updates of both the acting and the teaching value function, then the point of transition from acting to teaching is irrelevant to the learning progress of the two policies. Reducing the impact of the exploration policy to the learning updates allows for smoother interaction between the two policies and ensures us that we continue to learn the same policies. In principle, a Q-Teaching agent is able to update both its acting and teaching value functions continually and refine not only \textit{when} it should advise but also \textit{what} it should advise. 

\extVersion{In principle, Q-Teaching is able to continually update both its acting and teaching value functions. This is desirable since a teacher, while it is an alone actor, can already start learning an approximation of a teaching policy. To see how learning a teaching policy is feasible without having a student, we recall the previous section where we draw parallels of the teaching problem to the exploitation-exploration one. The insight is that since we think of \quot{not advising} as an exploration action we can evaluate the cost of not advising (i.e. learning a teaching policy) by estimating the exploration-exploitation regret as defined in the previous chapter. The teacher may not have a student, but it could still estimate how costly, performance-wise, is to explore.  

Likewise, when the teacher agent has a student, it can continue improving its acting policy (the one it learned while it was actually acting) that is, it can still update its acting policy based on the student's experiences and rewards which it observes. 
}

Since our goal is to introduce Q-Teaching as a flexible and generic enough method to be applied to multiple domains, we propose a series of state features for the teaching policy that we think are necessary. From our experiments, Q-Teaching works best with an augmented version of the acting task state space (see Table \ref{state}) similar to that of~\cite{zimmer} \rev{(Zimmer's method)}\AF{\#26}. Also in Table \ref{state}, note the role of the student's progress feature ($f3$): it homogenises the student's Markov chain by inducing a state feature for time (see Section \ref{sec:learningtoteach}).

\begin{table}[!h]
\caption{The augmented state feature set for the Teaching Task\AFold{After moving all,is this table here ok? Removing comments it goes upper}} 
\centering
\scalebox{0.8}{
\begin{tabular}{|l|c|p{5cm}|}
 \hline
a/a &	\textbf{Feature}&\textbf{Description} \\ \hline
 $f_1$ &Advice	&Binary feature indicating if the current state is the result of advice	\\
 $f_2$ &Budget&Remaining advice budget\\
$f_3$  &Student's progress	 &At least one informative feature for the student's learning progress (e.g., current episode)\\
$f_4$  &Student's intended action	 &The action announced by the student (optional)\\
$f_i$  & \vdots&Acting task original state features\\
 \hline
\end{tabular}
}
\label{state}
\end{table}

\subsection{Experiments and Results}

In this section, we present results from using Q-Teaching in the Pac-Man Domain. We evaluate both on-student's policy Q-Teaching and off-student's policy Q-Teaching, in two variations each: known or unknown student's intended action. Note that methods like Zimmer's and Mistake Correcting require knowledge of the student's intended action. 


\begin{table*}
	 \caption{Average student score in 1000 training episodes with teachers knowing the student's intended actions. The curves are averaged over 30 trials and the legend is ordered by score. The error bars represent the 95\% confidence intervals (CI) of the means. Non-overlapping CIs indicate statistically significant differences of the means whereas overlapping CIs are inconclusive}
	\begin{subfigure}{.5\linewidth}
	\centering
	\caption{Low-asymptote Sarsa student}
	\includegraphics[width=0.95\textwidth]{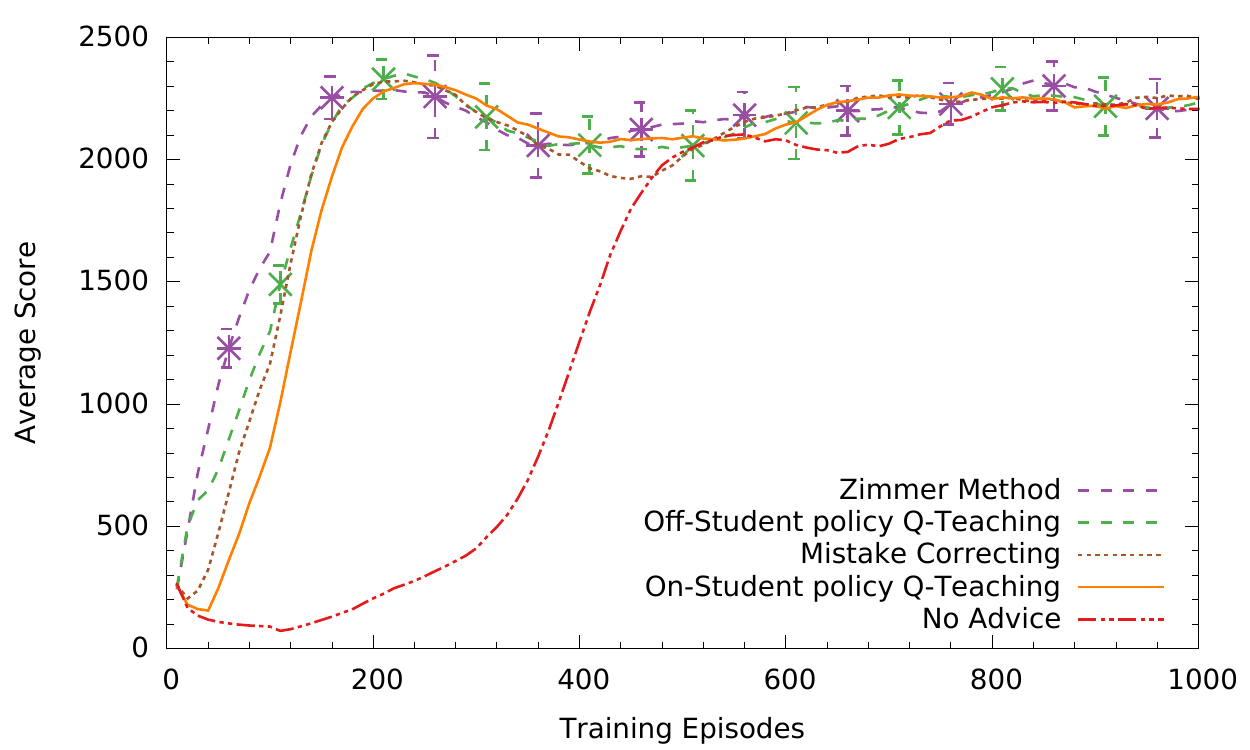}
	\label{fig:DepthS_student}
	\end{subfigure}%
	\begin{subfigure}{.5\linewidth}
	\centering
	\caption{High-asymptote Sarsa student}
	\includegraphics[width=0.95\textwidth]{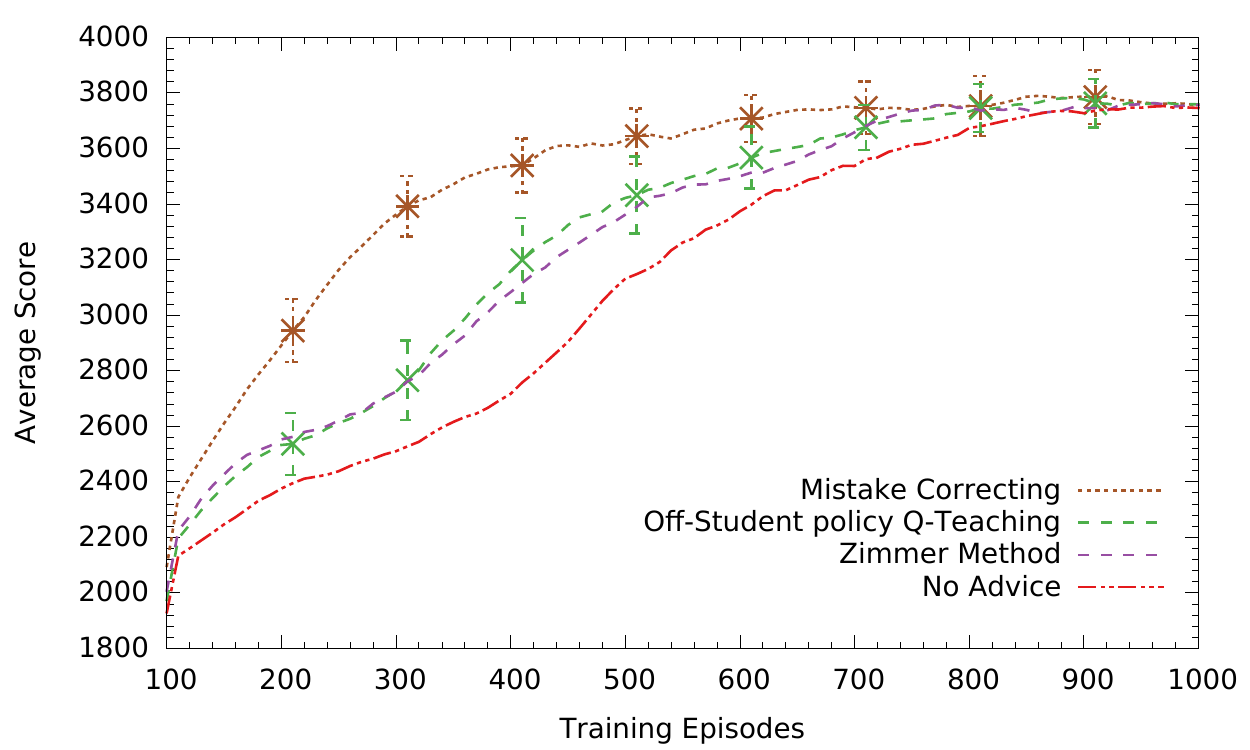}
	\label{fig:CustomS_student}
	\end{subfigure}

 \caption{Average student score of 1000 training episodes, averaged over 30 trials and ordered by total reward.\AF{\#6 should check and fill the exact training times}}
\begin{subtable}{.5\linewidth}

\centering
  \caption{Low-asymptote Sarsa student}
\scalebox{0.8}{
\begin{tabular}{|l|c|c|c|c|}
 \hline
&\parbox[t]{1.5cm}{\centering{\textbf{Student\\Intention}}\vspace{0.1cm}}&	\parbox[t]{1cm}{\centering{\textbf{Average\\Episode}}\vspace{0.1cm}} $\blacktriangledown$ &\parbox[t]{1.5cm}{\centering{\textbf{\rev{Avg. Session Time}}}\vspace{0.1cm}}&\parbox[t]{1.5cm}{\centering{\textbf{\rev{Avg. Budget Finish}}}\vspace{0.1cm}}\\ \hline
Zimmer method&Known&2145.35&1000 ep. &17 ep.\\
Off-Student Q-Teaching&Known&2103.28&42 ep.&31 ep.\\
Mistake Correcting&Known&2086.03&-&20 ep.\\
On-Student Q-Teaching&Known&2062.3&25 ep.&4 ep.\\
Off-Student Q-Teaching&Not Known&2048.47&18 ep.&4 ep.\\
Importance Advising&Not Known&2026.62&-&10 ep. \\
Early Advising&Not Known&1910.16&-&2 ep.\\
No Advice&-&1485.01&-&-\\
 \hline
\end{tabular}
}

\end{subtable}%
\begin{subtable}{.5\linewidth}

\centering
  \caption{High-asymptote Sarsa student}
\scalebox{0.8}{
\begin{tabular}{|l|c|c|c|c|}
 \hline
&\parbox[t]{1.5cm}{\centering{\textbf{Student\\Intention}}\vspace{0.1cm}}&	\parbox[t]{1cm}{\centering{\textbf{Average\\Episode}}\vspace{0.1cm}} $\blacktriangledown$ &\parbox[t]{1.5cm}{\centering{\textbf{\rev{Avg. Session Time}}}\vspace{0.1cm}}&\parbox[t]{1.5cm}{\centering{\textbf{\rev{Avg. Budget Finish}}}\vspace{0.1cm}}\\ \hline
Mistake Correcting&Known&3429.97&-&240 ep.\\
Off-Student Q-Teaching&Known&3235.12&28 ep.&13 ep.\\
On-Student Q-Teaching&Known&3221.75&22 ep.&11 ep.\\
Zimmer method&Known&3218.05&1000 ep. &8 ep.\\
Importance Advising&Not Known&3162.28&-&30 ep.\\
Off-Student Q-Teaching&Not Known&3116.09&23 ep.&12 ep.\\
Early Advising&Not Known&3112.84&-&2 ep.\\
No Advice&-&3079.66&-&-\\
 \hline
\end{tabular}
}

\end{subtable}

\label{perf}
\end{table*} 

We use two versions of students for the experiments. A low-asymptote and a high-asymptote Sarsa students. Referring to~\cite{torrey13} and Section \ref{pacman}, the low asymptote students receive a state vector of 16 primitive features related to the current game state while the high asymptote students receive a state vector of 7 highly engineered features providing more information. The low-asymptote students have significantly worse performance than the high-asymptote ones.

Additionally, we choose to bootstrap all compared teaching methods with the same acting policy in order to compare only their advice distribution performance and not their quality of the advice. The acting policy used for producing advice comes from a high-asymptote Q-Teaching agent after 1000 episodes of learning. Moreover, we use Sarsa students in order to emphasize the ability to advise students that are different to the teacher. All learning methods (Zimmer's and Q-Teaching) were trained for 500 teaching episodes (sessions) to be equally compared for their learning efficiency too.

The Q-Teaching learning parameters for the teaching policy were $\alpha=0.002$, decaying $\epsilon=0.5$ and $\lambda=0.9$ whereas all Sarsa students had $a=0.001$, \rev{$\epsilon=0.05$} and $\lambda=0.9$.

\rev{The evaluation was based on the student performance (game score) and using the \emph{Total Reward} TL metric \cite{JMLR09-taylor} divided by the fixed number of training episodes. The student performance is evaluated every 10 advising episodes (learning) for 30 episodes of acting alone (and not learning). For the comparisons between average score performances we used pairwise t-tests with Bonferroni correction. Statistically significant results are denoted with their significance level and they always refer to paired comparisons.}\AF{\#12\#1}

In Figure \ref{fig:DepthS_student}, teacher agents advise a low-asymptote Sarsa student who always announces its intended action. We can see Zimmer's method performs best and off-student's policy Q-Teaching comes second with a statistically significant difference ($p<0.05$). The heuristic based-method Mistake Correcting with a tuned threshold value of $t=100$  comes third. On-student's policy Q-Teaching performed worse than the previous three methods by a small margin, having not found an as good advice distribution policy (non-significant difference to Mistake Correcting). Finally, all methods performed statistically significantly better ($p<0.05$) than not advising, effectively speeding up the learning progress of the student.

In Figure \ref{fig:CustomS_student}, the teachers advise a high-asymptote Sarsa student. Here, the tuned version of Mistake Correcting ($t=200$) performed statistically significantly better ($p<0.05$) than all methods, with Q-Teaching methods coming second and third (respectively) and Zimmer's method coming next (having non significant differences between them). \remove{The high-asymptote Sarsa student is a fast learner and, similar to the conclusions in~\cite{torrey13}, needs all the advice early and not spread across many episodes. Both learning methods failed to learn a policy that provides the advice quickly enough.}

For the case when the teacher agent is not aware of the student's intended action, in Figure \ref{fig:DepthS_student_agnostic} the off-student-policy Q-Teaching performs best while Importance Advising ($t=200$) follows with a small performance difference (n.s.). Early Advising (giving all $B$ advice in the first $B$ steps) performs statistically significantly worse (at $p<0.05$) than both Q-Teaching and Importance Advising. In these experiments, we did not use on-student's policy Q-Teaching since that requires knowing the student's intended action to compute the reward.

In Figure \ref{fig:CustomS_student_agnostic}, advising a high asymptote Sarsa student, Q-Teaching had the second best performance with the heuristic-based method importance advising ($t=200$) performing better (non significant). For high performing students a poorly distributed advice budget can be much less effective. For example, if the teacher knows the student's intended action it does not spend advice in states where the student would anyway choose the correct action. This fact is emphasized in this specific case, since no advising did not perform significantly worse compared to the rest of the methods.

Finally, in Table \ref{perf} 	we can see the average total reward in 1000 training episodes for all the teaching methods. All methods knowing the student's intention performed better than those not, taking advantage of that knowledge. 

It is important to note that Q-Teaching, the only learning AuB method allowing students to not announce their intended action, performed relatively well compared to methods that know the student's intended action, which is an advantage of the proposed method. 

Another advantage is that off-student's policy Q-Teaching can use the same teaching policy for very different students since it is not directly influenced from the student's policy and the rewards received by the student when not advising (such as in the Zimmer method). This is a significant advantage in terms of learning speed and versatility since heuristic methods have to be manually tuned for each student separately to find the optimum threshold, $t$. 

Moreover, while Zimmer and Q-Teaching methods were both trained for 500 episodes (sessions), Q-Teaching training completed significantly faster since the Zimmer method has to observe all 1000 episodes of \rev{each student session}\AF{\#6} to complete just one of its own, whereas Q-Teaching has an upper bound for its episode completion. \rev{This upper bound is} the algorithmic convergence of the student (e.g, the low-asymptote student requires only 500 episodes to converge) and in most cases it will complete much faster, when the budget finishes (around the 30th episode for the low-asymptote student). \rev{More specifically, in Table \ref{perf} we can see the average training time needed for each teacher in terms of the average observed \emph{student} episodes in each of the 500 teacher episodes. In general, our proposed methods need at least $\times25$ less training time than the Zimmer's method. We should also note here that although non-learning methods do not need training time they require a significant and variable amount of manual parameter tuning to achieve the reported performance.} \AF{\#6}

On-student policy Q-Teaching did not perform as well expected, the main problem being the non-stationary reward depending on the student's changing policy. We believe that this method needs significantly more training time than the off-student's policy Q-Teaching because of its non-stationary reward and it probably needs more informative features for the student's current status. In our case, this was only its training episode which is the most basic information available for the student. Moreover, the training episode feature is student-dependent since its meaning varies among students---some students learn faster than others.

\section{Related Work}
\label{sec:related}

There are several types of related work in the area of helping to learn. Some of this work focuses on teaching in non-RL settings~\cite{chakraborty2006teaching,stone2010ad}.

In the field of \emph{transfer learning} in RL~\cite{taylor2009transfer}, an agent uses knowledge from a source task to aid its learning in a target task. However, agents perform transfer knowledge from one task to another and in an off-line manner. Other differences of this typical TL setting to Agent Advising are described in section \ref{background2} of this article.

\AFold{In this par. you suggested to define the contrasts (?) is it ok now?}More closely related work has one RL agent teach another without a direct knowledge transfer. Examples of such works include \emph{imitation learning}~\cite{lin1992self} and \emph{apprentice learning}~\cite{Clouse:1996:IAL:923832}. In these approaches an expert provides demonstrations of the task to a student, then the student has to extract a policy by either learning directly from them or building a model to generate mental experience. In our setting, the teacher does not provide a full-policy trajectory and has a limitation on the number of interventions (advice budget). Moreover, we do not require a student with special processing abilities except that of being able to receive advice   

In~\cite{torrey13} a non-learning teaching framework for RL tasks is proposed based on action advice. The methods presented there are described in more detail in section \ref{background2}. One drawback of these methods is that since they are based solely on the teacher's q-values they are not able to handle non-stationarity in the student’s learning task, and also have to be given a threshold of q-value differences, above which a state is considered important. This parameter needs to be manually tuned for each student in contrast to off-student's policy Q-Teaching which can learn a more generic teaching policy focusing on the criticalities of the state space. 

Also, since the methods presented in~\cite{torrey13} are heuristic-based and not based on adaptive learning, the agent may spent all of its advising budget on early learning steps of the student that satisfy the importance threshold, while it may later experience even more important states that further exceed the given threshold. \extVersion{The above method merges with another proposed method in~\cite{torrey13}, predictive advising, which is able to predict a student's action and therefore choose more correctly the appropriate time to advise. This is a batch method of learning a new model of student behaviour after every episode, so it can not be considered as efficient, also a student not following an advice or a student exploring are considered as data noise thus not formulating stochasticity in a more natural way as our proposed method.}

The only other learning method for advising is introduced in~\cite{zimmer} (Zimmer's method). The method proposed there is described in more detail in section \ref{background2}. One significant difference is that the method is based on the same reward received by the student, needing ad-hoc modifications for each task to encourage teacher towards a better advising policy. Our method uses a domain-independent reward signal based on the acting task q-values and can be directly used in any task. Moreover, their method has greater data complexity since a complete batch of student training episodes is required for just one training episode of the teacher. As discussed in the previous section, our method may finish one teaching episode as early as the budget finishes; that is multiple times faster completion of one episode. Finally but most important, Q-Teaching can be used in the more realistic setting where there is no knowledge of the student's intended action.

Concerning the model selection criteria proposed in Section \ref{sec:modelselection} for the teacher's acting policy, to the best of the author's knowledge there is no other work in the relevant literature examining these criteria and furthermore proposing performance variance, and specifically CV, as an important one. Most relevant works choose models based on their average performance, which as discussed previously, is not enough to evaluate the teaching effectiveness of a policy sampled infrequently and in parts.

\section{Conclusions and Future Work}
\label{sec:conclusions}
In this article, we discussed and proposed criteria, considerations and methods for the problem of learning teaching policies to produce and distribute advice. 

Concerning advice production, we identify a model selection problem for the teacher, selecting the appropriate acting policy from which to advise. The experiments showed the significant relation of CV to the teaching performance, promoting CV as an important criterion---among others tested---for selecting acting policies for advising. \rev{Moreover, average-reward RL was found to produce effective policies for sparse advising under budget, although these policies may under-perform when used as acting ones.

Concerning advice distribution (i.e., teaching policy)
we proposed a novel \rev{representation} of the learning to teach problem as a \rev{constrained exploitation reinforcement learning problem}\AF{\#1,\#15 under-stated it}. Based on this \rev{representation }\AF{\#1,\#15 replaced the word formulation} we proposed a novel RL algorithm for learning a teaching policy, Q-Teaching, able to advise even when not having knowledge of the student's intended action. Q-Teaching was found to perform at least equally well with other compared methods while needing significantly less training time.}\AF{\#40 Matt, are the last 4 lines ok...?}

\remove{\rev{Q-Teaching uses convergence horizon (see Definition \ref{converge}) as one of its episode stopping criteria along with advice budget. One limitation of convergence horizon as defined here is that in tasks where rewards are very sparse, the algorithm might converge (thus, the teaching session) without ever receiving one. In principal, Q-Teaching could use an improved stopping criteria that will take reward sparsity into account too.}\AF{\#19 1st response version}}


Advice distribution under budget is a challenging problem, both theoretically and practically, posing a series of problems such as the non-stationarity of the teaching task, as a result of having a learning student as part of the environment. Efficient and principled handling of the budget constraint is another challenge. \remove{Moreover, the constrained exploitation problem in RL and its connection with the advising problem should be studied further.}

From our experiments, Q-Teaching can be considered a promising method based on a more formal understanding of the problem. It is significantly more efficient in terms of data complexity than Zimmer's method, and it can learn teaching policies without the assumption of having knowledge for the student's intended action.

There are several future directions. Q-Teaching could be adapted to student agents with specific \quot{disabilities} and could also be tested under different budget costraints to examine how budget affects its teaching policies. \AF{\#13} \rev{Also, off-student Q-Teaching could be tested on multi-student scenarios since not fitting to a particular student could be proven effective when teaching multiple different students}\AF{\#38}. Moreover, the theoretical properties of the algorithms should be studied, especially the case of learning a teaching and an acting policy at the same time, e.g., under which specific assumptions a teaching policy converges. 

\rev{The general usefulness of CV as a criteria for selecting teachers should be studied. Specifically, how teacher selection criteria such as CV are capturing the robustness of a policy when that policy is used sparingly for advising.} \AF{\#7}

Finally, other teaching architectures and representations should be studied, allowing, for example, a teacher to use only one value function for both advising under a budget and acting. Such a hybrid agent transitions smoothly from its actor role to the teacher's one. A unified architecture and knowledge representation would further reveal the deep connection between acting and teaching, one we strongly believe exists.

\AFold{Where in the text could I cite the IJCAI-16 paper? You mean instead of citing the one with just Yusen and you (Online TL) or both? Also, can you please give me its bib entry? the one I can get is from the arxiv version}


%
%
%

\bibliography{transferbib}
\bibliographystyle{plain}

\end{document}